\documentclass[10pt,twocolumn,letterpaper]{article}

\usepackage{cvpr}
\usepackage{times}
\usepackage{epsfig}
\usepackage{graphicx}
\usepackage{amsmath}
\usepackage{amssymb}
\usepackage{verbatim}
\usepackage{xcolor}
\usepackage{multirow}
\usepackage{bbding}
\usepackage{pifont}
\usepackage{makecell}
\usepackage{wasysym}

\usepackage[pagebackref=true,breaklinks=true,letterpaper=true,colorlinks,bookmarks=false]{hyperref}

 \cvprfinalcopy % *** Uncomment this line for the final submission

\def\cvprPaperID{58} % *** Enter the CVPR Paper ID here

\def\parsp{\vspace{.04in}}

\def\regularcnn{large-scale classifier\ }
\ifcvprfinal\fi
\begin{document}

%%%%%%%%% TITLE
\title{Few-Shot Open-Set Recognition using Meta-Learning}

\author{Bo Liu\\
UC, San Diego\\
{\tt\small boliu@ucsd.edu}
\and
Hao Kang\\
Wormpex AI Research\\
{\tt\small haokheseri@gmail.com}
\and
Haoxiang Li\\
Wormpex AI Research\\
{\tt\small lhxustcer@gmail.com}
\and
Gang Hua\\
Wormpex AI Research\\
{\tt\small ganghua@gmail.com}
\and
Nuno Vasconcelos\\
UC, San Diego\\
{\tt\small nuno@ece.ucsd.edu}
}

\maketitle
%\thispagestyle{empty}

%%%%%%%%% ABSTRACT
\begin{abstract}
    The problem of open-set recognition is considered. While previous approaches only consider this problem in the context of large-scale classifier training, we seek a unified solution for this and the low-shot classification setting. It is argued that the classic softmax classifier is a poor solution for open-set recognition, since it tends to overfit on the training classes. Randomization is then proposed as a solution to this problem. This suggests the use of meta-learning techniques, commonly used for few-shot classification, for the solution of open-set recognition. A new {\it oPen sEt mEta LEaRning\/} (PEELER) algorithm is then introduced. This combines the random selection of a set of novel classes per episode, a loss that maximizes the posterior entropy for examples of those classes, and a new metric learning formulation based on the Mahalanobis distance. Experimental results show that
    PEELER achieves state of the art open set recognition
    performance for both few-shot and large-scale recognition.
    On CIFAR and miniImageNet, it achieves substantial gains in seen/unseen class detection AUROC for a given seen-class classification accuracy.

\end{abstract}

%%%%%%%%% BODY TEXT
\section{Introduction}

The introduction of deep convolutional neural networks (CNNs) has catalyzed large advances in computer vision.
Most of these advances can be traced back to advances on object recognition, due to the introduction of large scale datasets, such as ImageNet~\cite{deng2009imagenet}, containing many classes and many examples per class. Many modern computer vision architectures are entirely or partially based on recognition networks. In the large scale setting, 
CNN-based classifiers trained by cross-entropy loss and mini-batch SGD have excellent recognition performance, achieving state of the art results on most recognition benchmarks.
%CNN-based softmax classifiers trained by cross-entropy have excellent recognition performance, achieving state of the art results on most recognition benchmarks. 

% \begin{table}[t]
%     % \footnotesize
% \begin{small}
%   \centering
%   \begin{tabular}{|c|c|c|}
%     \hline
%     \thead{Recognition} & \thead{Training} & \thead{Testing} \\
%     %\hline
%     \makecell{Typical \\ Paradigms} & \makecell{No. of Samples \\ per Class} & \makecell{Supports \\ unseen Class?} \\
%     \hline
%     Closed-set~\cite{krizhevsky2014cifar} & Large & No \\
%     \hline
%     Few-shot~\cite{Ravi2017,Finn2017,Vinyals2016, Snell2017} & Small & No \\
%     \hline
%     Open-set~\cite{Ge2019,Bendale2016} & Large & Yes \\
%     \hline
%     Few-Shot Open-set & Small & Yes  \\
%     \hline
%   \end{tabular}
%   \caption{Comparison between different tasks.}
%   \label{tab:intro}
% \end{small}
%   \vspace{-2em}
% \end{table}

\begin{table}[t]
     \footnotesize
  \centering
  \begin{tabular}{|c|c|c|}
    \hline
    \thead{\makecell{Typical \\ Recognition Paradigms}} & \thead{\makecell{No. of Samples \\ per Training Class}} & \thead{\makecell{Supports unseen \\ Class in Testing?}}\\
    \hline
    Closed-set~\cite{krizhevsky2014cifar} & Large & No \\
    \hline
    Few-shot~\cite{Ravi2017,Finn2017,Vinyals2016, Snell2017} & Small & No \\
    \hline
    Open-set~\cite{Ge2019,Bendale2016} & Large & Yes \\
    \hline
    Few-Shot Open-set & Small & Yes  \\
    \hline
  \end{tabular}
  \vspace{0.5em}
  \caption{Comparison between different recognition tasks.}
  \label{tab:intro}
  \vspace{-1.5em}
\end{table}

However, variations of the recognition setting can lead to substantial performance degradation. Well known examples include few-shot learning~\cite{Ravi2017, Finn2017, Vinyals2016, Snell2017}, where only a few training examples are available per class, domain adaptation~\cite{tzeng2017adversarial}, where a classifier trained on a source image domain (e.g. synthetic images) must be deployed to a target domain (e.g. natural images) whose statistics differ from those of the source, long tailed recognition~\cite{Liu2019}, where the number of examples per class is highly unbalanced, or problems with label noise~\cite{natarajan2013learning, sukhbaatar2014learning}. All these {\it alternative recognition settings\/} stress the robustness, or ability to generalize, of the \regularcnn.
%and many alternative architectures have been proposed to address them. 

More recently, there has been interest in endowing CNNs with self-awareness capabilities. At the core, self-awareness implies that, like humans, CNNs should identify what they can do and refuse what they cannot. Several variants of this problem have been proposed, including out-of-distribution detection~\cite{Hendrycks2016}, where the CNN rejects examples outside the training distribution (e.g. images from other domains or adversarial attacks~\cite{huang2017adversarial}), realistic classification~\cite{lu2017safetynet}, where it rejects examples that it deems too hard to classify, or open-set recognition~\cite{Bendale2016}, where the examples to reject are from novel classes, unseen during training. While several techniques have been proposed, a popular approach is to force the CNN to produce high entropy posterior distributions in the rejection regions. Rejections can then be identified as examples that give rise to such distributions.

In this work, we consider open-set recognition. This has been mostly addressed in the large-scale setting, using solutions based on the \regularcnn. These attempt to recognize the novel classes by post-processing the posterior class distribution~\cite{Bendale2016}, defining a ``reject'' class trained with examples that are either artificially generated~\cite{Ge2019} or sampled from the training classes~\cite{Schlachter2019}, or a combination of the two. 
%Such rejection strategies have been shown successful for classifiers trained in the large-scale setting.
%All these methods have been developed and tested exclusively in the large-scale recognition setting.
We seek to generalize  open-set recognition beyond the large-scale setting, proposing that it should target a continuum of tasks that includes all the alternative recognition settings. Since training over many settings can be complex, we consider the two at the extremes of this continuum: large-scale and few-shot recognition. 

There are three main reasons for this. First, open-set recognition is a challenge under all settings. A recognizer trained in the few-shot regime is not less likely to face unseen classes. An open-set recognition technique that also supports the few-shot setting is thus more useful than the one that does not. Second, few-shot open-set recognition is harder to solve than large-scale open-set recognition, due to the scarcity of labeled data. Hence, the few-shot setting poses a greater challenge to open-set recognition research. Third, like open-set recognition, the main challenge of few-shot recognition is to make accurate decisions for data unseen during training. Since this makes robustness the main trait of few-shot architectures, these architectures are likely to also excel at open-set recognition. Most notably, they are likely to beat the large-scale classifier, which tends not to score highly along the robustness dimension. 

The main limitation of the \regularcnn (trained by cross-entropy) is that it tends to overfit the training classes. Because the ideal embedding for classification maps all examples of each class into a unique point in feature space, the \regularcnn tends to learn embeddings that are only locally accurate. While the metric structure of the embedding is reflected by the semantic distances in the neighborhood of class representatives (the parameter vectors of the softmax layer), these distances are meaningless away from the latter. In result, \regularcnn embeddings underperform on tasks that require generalization beyond the training set, such as image retrieval~\cite{gordo2016deep}, face identification~\cite{liu2017sphereface}, pose invariant recognition~\cite{ho2019pies}, person re-identification~\cite{zhong2017re}, or few-shot learning~\cite{Vinyals2016}. Embeddings derived from the metric learning literature~\cite{ZhirongWuYuanjunXiongStellaX.Yu2018} or explicitly designed for problems like few-shot recognition~\cite{wang2018low} tend to capture the metric structure of the data more broadly throughout the feature space and achieve better performance on these generalization critical tasks. It is thus expected that these embeddings will outperform the \regularcnn on the open set recognition problem.

In this work, we investigate the open-set performance of a class of popular solutions to the few-shot problem, known as meta-learning (ML)~\cite{Vinyals2016}. ML approaches replace the traditional mini-batch CNN training by episodic training. Episodes are generated by randomly sampling a subset of the classes and a subset of examples per class, to produce a support and query set to which a classification loss is applied. This randomizes the classification task for which the embedding is optimized at each ML step, producing a more robust embedding, less overfitted to any particular set of classes. We generalize the idea to open-set recognition, by randomly selecting a set of novel classes per episode, and introducing a loss that maximizes the posterior entropy for examples of these classes. This forces the embedding to better account for unseen classes, producing high-entropy posterior distributions beyond the regions of the target classes. 

This solution has at least two strong benefits. First, because it draws on a state of the art approach to few shot learning, it immediately extends open set recognition to few-shot problems. Second, because ML embeddings are robust and open set recognition is about generalization, the performance of the latter improves {\it even in the large-scale setting\/}. 
This is shown by extensive experiments with various open-set recognition benchmarks, where we demonstrate significant improvements over the state of the art. We also investigate the role of the metric used in the feature space on open-set recognition performance, showing that a particular form of the Mahalanobis distance enables significant gains over the commonly used Euclidean distance.

Overall, the contributions of this work can be summarized as follows:
\begin{itemize}
    \item A new ML-based formulation of open-set recognition. This generalizes open-set to the {\it few-shot} recognition setting. 
    \item A new episodic training procedure, combining the cross-entropy loss and a novel {\it open-set loss} to improve open-set performance on both the large-scale and few-shot settings.
    \item A {\it Gaussian embedding} for ML-based open-set recognition.
\end{itemize}

% \begin{figure*}[t!]
% \centering{
% 		\begin{tabular}{cc} 
% 		\includegraphics[width=0.47\textwidth]{figs/regular_model.pdf} & \includegraphics[width=0.47\textwidth]{figs/fewshot_model.pdf}
% 		\end{tabular}}
% 	\caption{Our framework on Regular and Few-shot Open-Set Recognition: we assume Gaussian distribution of each class in feature space and introduce an entropy loss in training the networks.}
% 	\label{fig:model}
% \end{figure*}

%-------------------------------------------------------------------------
\section{Related Work}
\parsp
\noindent{\bf Open-Set Recognition:}
Open-set recognition addresses the classification setting where inference can face samples from classes unseen during training.
%In traditional recognition problem, we assume all classes during testing are also in training set. We name this common assumption as closed-set recognition. On the contrary, open-set recognition means new classes unseen in training appear in testing. 
The goal is to endow the open-set classifier with a mechanism to reject such samples. One of the first deep learning approaches was the work of Scheirer et al.~\cite{Bendale2016}, which proposed an extreme value parameter redistribution method for the logits generated by the classifier. 
%They first fit activations in the penultimate network layer into a Weibull distribution per class. Then with libMR models, the logits are rebalanced with an extra ``class'', which is used to reject unknown classes. Although a probability method is applied to give the rejection ability to SoftMax classifier, this work doesn't train the whole model, and as a result, the ability to detect unseen classes is limited.
Later works considered the problem in either discriminative or generative models. Schlachter et al.~\cite{Schlachter2019} proposed an intra-class splitting method, where a closed-set classifier is used to split data into typical and atypical subsets, reformulating open-set recognition as a traditional classification problem. G-OpenMax~\cite{Ge2019} utilizes a generator trained to synthesize examples from an extra class that represents all unknown classes. 
%And both known samples and generated samples are used to train the Weibull model. 
Neal et al.~\cite{Neal2018} introduced counterfactual image generation, which aims to generate samples that cannot be classified into any of the seen classes, producing an extra class for classifier training.

All these methods reduce the set of unseen classes to one extra class. While open samples can be drawn from different categories and have significant visual differences, they assume that a feature extractor can map them all into a single feature space cluster. While theoretically possible, this is difficult. Instead, we allow a cluster per seen class and label samples that do not fall into these clusters as unseen. We believe this is a more natural way to detect unseen classes.

\parsp
\noindent{\bf Out-of-Distribution:} A similar problem to open-set recognition is to detect out-of-distribution (OOD) examples.
Typically~\cite{Hendrycks2016}, this is formulated as the detection of samples from a different dataset, {\it i.e.} a different distribution from that used to train the model. While~\cite{Hendrycks2016} solves this problem by directly using the softmax score, later works~\cite{Liang2017, Vyas2018} improve results by enhancing reliability. This problem differs from open-set recognition in that the OOD samples are not necessarily from unseen classes. For examples, they could be perturbed versions of samples from seen classes, as is common in the adversarial attack literature. The only constraint is that they do not belong to the training distribution. Frequently, these samples are easier to detect than samples from unseen classes. In the literature, they tend to be classes from other datasets or even images of noise. This is unlike open-set recognition, where unseen classes tend to come from the same dataset.

%the  could be from the seen classes we try to extend the fixed softmax classifier widely used in deep learning to detect open samples. In OOD problem, the model is trained to reject samples out of the learned distribution, and as a result, attempt to solve the problem where training and testing data are not consistent.

\parsp
\noindent{\bf Few-Shot Learning:}
Extensive research on few-shot learning~\cite{Ravi2017, Finn2017, Vinyals2016, Snell2017, Sung2018, Ravichandran2019, Gidaris2018, Oreshkin2018, Li2019,8935497} has emerged in recent years. These methods can be broadly divided into two branches: optimization and metric based.
Optimization based methods deal with the generalization problem by unrolling the back-propagation procedure. 
Specifically, Ravi et al.~\cite{Ravi2017} proposed a learner module that is trained to update to novel tasks. MAML~\cite{Finn2017} and its variants~\cite{Finn2018} proposed a training procedure, where the parameters are updated based on the loss calculated by secondary gradients. 
%And as a result, the model is trained to have the ability that can be easily adapted to a new task.
Metric based approaches attempt to compare feature similarity between support and query samples. 
%To achieve this, they managed to train a embedding function that can map both support and query inputs into a shared feature space. 
Vinyals et al.~\cite{Vinyals2016} introduced the concept of episode training, where the training procedure is designed to mimic the test phase, which was used with a cosine distance to train recurrent networks.  The prototypical network~\cite{Snell2017} introduced class prototypes constructed from support set features, by combining metric learning and cross-entropy loss. 
The relation network~\cite{Sung2018} explored the relationship between a pair of support and query features implicitly with a neural network, instead of building a metric directly on feature space.

Since a feature space metric is also useful for open-set recognition, we mainly focus on metric based approaches to few-shot learning. Although several methods have demonstrated promising results for the few-shot task, it is still unclear whether the resulting classifiers can successfully reject unseen samples. In this work, we explore this problem under both the large-scale and few-shot scenarios.

\parsp
\noindent{\bf Learning without Forgetting:} Several works~\cite{Gidaris2018, Qi2018, Qiao2018, Ren2018} extend traditional few-shot to learning without forgetting. The model is trained to work on an extra few-shot problem and maintains its performance on the original recognition problem. Our work will focus on the traditional few-shot problem, this direction is beyond our scope, and can be discussed in the future.

% \begin{table*}[t!]
% 	\begin{footnotesize}
% 		\centering
% 		\begin{tabular}{|c|c|c|c|c|}
% 		    \hline
% 		    \multicolumn{2}{|c|}{Task} & \multicolumn{2}{|c|}{Training} & \multirow{2}{*}{Testing (classes)} \\
% 		    \cline{0-3}
% 		    Closed vs. Open & Large vs. Low & number of samples & classes & \\
% 			\hline
% 			Closed & Large & large & $\mathbb{C}^s$ & $\mathbb{C}^s$ \\
% 			\hline
% 			Closed & Low & small & $\mathbb{C}^s$ & $\mathbb{C}^s$ \\
% 			\hline
% 			Open & Large & large & $\mathbb{C}^s$ & $\mathbb{C}^s$ and $\mathbb{C}^u$ \\
% 			\hline
% 			Open & Low & small & $\mathbb{C}^s$ & $\mathbb{C}^s$ and $\mathbb{C}^u$ \\
% 			\hline
% 		\end{tabular}
% 		\caption{Comparison between different tasks.
% 		\label{tab:tasks}}
% 	\end{footnotesize}
% \end{table*}

%-------------------------------------------------------------------------
\section{Classification tasks and meta Learning}~\label{sec:cls_setting}
In this section, we discuss different classification settings (Sec.~\ref{sec:setting}) and meta learning (Sect.~\ref{sec:ML}), which motivates and provides the foundation for the proposed solution to open-set recognition in both the large-scale and few-shot settings.
\subsection{Classification settings~\label{sec:setting}}
%In this section, we review the definition of different classification settings previously used in the literature.
%and introduce... 
%to clarify some definitions, we will first review the traditional open-set recognition problem, and recent develops in few-shot learning. And then, the new problem few-shot open-set recognition will be proposed.

\parsp
\noindent{\bf Softmax classification.} The most popular deep learning architecture for object recognition and image classification is the softmax classifier. This consists of an embedding that maps images $\mathbf{x} \in {\cal X}$ into feature vectors $f_\phi(\mathbf{x}) \in {\cal F}$, implemented by multiple neural network layers, and a softmax layer with linear mapping that estimates class posterior probabilities with 
\begin{equation}
    p(y = k | \mathbf{x}; \phi, \mathbf{w}_k) = \frac{\exp(\mathbf{w}_k^Tf_\phi(\mathbf{x}))}
    {\sum_{k'}\exp(\mathbf{w}_{k'}^Tf_\phi(\mathbf{x}))}
    \label{eq:softmax}
\end{equation}
where $\phi$ denotes all the embedding parameters and $\mathbf{w}_k$ is a set of classifier weight vectors. 
Recent works~\cite{Snell2017} have combined metric learning with softmax classification, implementing the softmax layer with a distance
function 
\begin{equation}
    p_\phi (y = k | \mathbf{x}) = \frac{\exp(-d(f_\phi(\mathbf{x}),\mathbf{\mu}_k))}{\sum_{k'}\exp(-d(f_\phi(\mathbf{x}),\mathbf{\mu}_{k'}))}
    \label{eq:softmaxdist}
\end{equation}
where $\mathbf{\mu}_k = E[f_\phi(\mathbf{x})|y=k]$~\cite{Snell2017}.
The softmax classifier is learned with a training set $\mathbb{S} = \{(x_i^s, y_i^s)\}_{i=1}^{n^s}$, where $y_i^s \in \mathbb{C}^s, \forall i$, $\mathbb{C}^s$ is a set of training image classes, and $n^s$ the number of training examples. This consists of finding the classifier and embedding parameters that minimize the cross entropy loss
\begin{equation}
    {\cal L}_{CE} = \sum_{(x_i^s, y_i^s)} -\log p(y_i^s | \mathbf{x}_i^s)
    \label{eq:softmaxce}
\end{equation}
Recognition performance is evaluated on a test set $\mathbb{T}=\{(x_i^t, y_i^t)\}_{i=1}^{n^t}$, where $y_i^t \in \mathbb{C}^t, \forall i$, $\mathbb{C}^t$ is a set of test classes, and $n^s$ the number of test examples.

\parsp
\noindent{\bf Closed vs. open set classification.} Under the traditional definition of classification, the sets of training and test classes are identical, i.e. $\mathbb{C}^s =  \mathbb{C}^t$. This is denoted as {\it closed set classification\/}. Recently, there has been interest in an alternative {\it open set\/} classification setting, where $\mathbb{C}^t = \mathbb{C}^s \cup \mathbb{C}^u$. In this case, the classes in $\mathbb{C}^s$ are denoted 
as {\it seen\/} (during training) classes and the classes in $\mathbb{C}^u$ as {\it unseen\/}.

\parsp
\noindent{\bf Large-scale vs few-shot recognition.} In the large-scale recognition setting, the number $n^s$ of training examples is quite high, reaching the millions for datasets like ImageNet. On the contrary, for few-shot recognition, this number is quite low, usually less than twenty examples per class. A few-shot problem with $K$ training examples per class is commonly known as $K$-shot recognition. Note that the number $n^t$ of test examples plays no role in differentiating these two settings. Since these examples are only used for performance evaluation, the test set should have the same cardinality under the two settings. As shown in Table~\ref{tab:intro}, different combinations of properties in the scale and coverage of the training data define different classification paradigms. The few-shot open-set setting is largely unexplored in the literature and the focus of this paper.

\subsection{Meta-learning~\label{sec:ML}} 
Meta-learning (ML) addresses the problem of ``learning to learn''. In this case, a meta-learner learns a learning algorithm by inspection of many learning problems. For this, the meta-learner relies on a meta training set $\mathbb{MS} = \{(\mathbb{S}^s_i, \mathbb{T}^s_i)\}_{i=1}^{N^s}$, where $(\mathbb{S}^s_i, \mathbb{T}^s_i)$ are the training and test set of the $i^{th}$ learning problem and $N^s$ the number of learning problems used for training; and a meta test set $\mathbb{MT} = \{(\mathbb{S}^t_i, \mathbb{T}^t_i)\}_{i=1}^{N^t}$, where $(\mathbb{S}^t_i, \mathbb{T}^t_i)$ are the training and test set of the $i^{th}$ test learning problem and $N^t$ is the number of learning problems used for testing. Given $\mathbb{MS}$, the meta-learner learns how to map a pair $(\mathbb{S}, \mathbb{T})$ into an algorithm that leverages $\mathbb{S}$ to optimally solve $\mathbb{T}$. 
%For example, an algorithm, that maps $\mathbb{S}$ into an estimate of the parameters, updates for the softmax classifier that the application of back-propagation to $\mathbf{S}$ would produce~\cite{Finn2017}.

The procedures are as follows. At meta iteration $i$, a meta-model $h$ is initialized with the one produced by the previous meta-iteration. Two steps are then performed. First, the meta-learning algorithm performs the mapping
\begin{equation}
    h' = {\cal M}(h,\mathbb{S}^s_i)
    \label{eq:ML1}
\end{equation}
to produce an estimate $h'$ of the optimal model for training set $\mathbb{S}^s_i$. The test set $\mathbb{T}^s_i$ is then used to find the model
\begin{equation}
    h^* = \arg \min_{h} \sum_{(x_k, y_k) \in \mathbb{T}^s_i} L[y_k, h'(x_k)]
    \label{eq:ML2}
\end{equation}
for a suitable loss function $L$, e.g. cross-entropy, using a suitable optimization procedure, e.g. back-propagation. The resulting $h^*$ is finally returned as the optimal meta-model for meta-iteration $i$.
During testing, the final meta-model $h^*$ and a training set $\mathbb{S}^t_i$ from the meta test set $\mathbb{MT}$ 
are used by the meta-learner to produce a new model
\begin{equation}
    h'' = {\cal M}(h^*,\mathbb{S}^t_i)
\end{equation}
whose performance is evaluated with $\mathbb{T}^t_i$.

\parsp
\noindent{\bf ML for few-shot recognition.} While different approaches have been proposed to apply ML to few shot recognition, in this work we adopt the procedure introduced by the popular prototypical network architecture~\cite{Snell2017}, which is the foundation of various other approaches. In this context, ML is mostly a randomization procedure. The pairs $(\mathbb{S}_i, \mathbb{T}_i)$ are denoted {\it episodes\/}, the training sets $\mathbb{S}_i$ are denoted {\it support sets} and the test sets $\mathbb{T}_i$ {\it query sets\/}. 
%The classifier is first initialized by training on an {\it auxiliary\/} set of classes, which are not part of the low-shot problem, using the large-scale setting. 
The meta-training set $\mathbb{MS}$ is generated by sampling train and test classes. In particular, the training set $\mathbb{S}^s_i$ of the $i^{th}$ episode is obtained by sampling $N$ classes from the set of classes of the low-shot problem, and $K$-examples per class. This defines a set of $K$-shot learning problems and is known as the {\it $N$-way $K$ shot problem}. The model $h$ is the softmax classifier of~(\ref{eq:softmaxdist}), the meta-learning mapping of \eqref{eq:ML1} implements the Gaussian mean maximum likelihood estimator
\begin{equation} \label{eq:proto}
    \mathbf{\mu}_k = \frac{1}{|\mathcal{P}_{i,k}|}\sum_{\mathbf{x}_j \in \mathcal{P}_{i,k}}f_\phi (\mathbf{x}_j)
\end{equation}
where $\mathcal{P}_{i,k}=\{\mathbf{x}_j \in \mathbb{S}^s_i | y_j = k)$ is the set of support samples from class $k$,  %$\mathbb{S}^s_i$ to update the parameters $\mathbf{w}_k$ 
and back-propagation is used in~(\ref{eq:ML2}) to update the embedding $f_\phi$.
%\textcolor{blue}{The embedding is finally trained by a comparison between support set $\mathbb{S}^s_i$, and query set $\mathbb{T}^s_i$, typically using the cross-entropy loss, back propagation on the query set $\mathbb{T}^s_i$.}
% The embedding is finally trained on the support set $\mathbb{S}^s_i$, typically using the cross-entropy loss, back propagation, and the embedding learned in the previous episode as initialization. The query set $\mathbb{T}^s_i$ is used to optimize hyper-parameters [IS THIS RIGHT? WHAT IS THE TEST SET USED FOR?].

\parsp
\noindent{\bf Benefits of ML for open-set classification.}
The ML procedure is quite similar to the standard mini-batch procedure for training softmax classifiers. 
Aside from the use of episode training instead of the more popular mini-batch training, there are two main differences. 
First, $f_\phi$ is updated by back-propagation from examples in $\mathbb{T}^s_i$ rather than $\mathbb{S}^s_i$. This is advantageous for few-shot learning due to the consistency with meta-testing, which is not the case in the large-scale regime. 
Second, mini-batches of random examples from all classes are replaced by examples from the subset of $N$ classes contained in $\mathbb{S}^s_i$ and $\mathbb{T}^s_i$. Randomizing the classification task learned per episode forces the embedding $f_\phi$ to generalize better to unseen data. This property makes ML a good solution for few-shot learning, where the training data lacks a good coverage of intra-class variations. In this case, large subsets of test samples from the known classes are unseen during training. We propose that the same property makes ML a good candidate for open-set classification, where by definition the classifier must deal with unseen samples from unseen classes. This observation motivates us to design a unified ML-based solution to open-set classification that supports both large-scale and few-shot classification.

\begin{figure}[t!]
\centering{
	\includegraphics[width=0.45\textwidth]{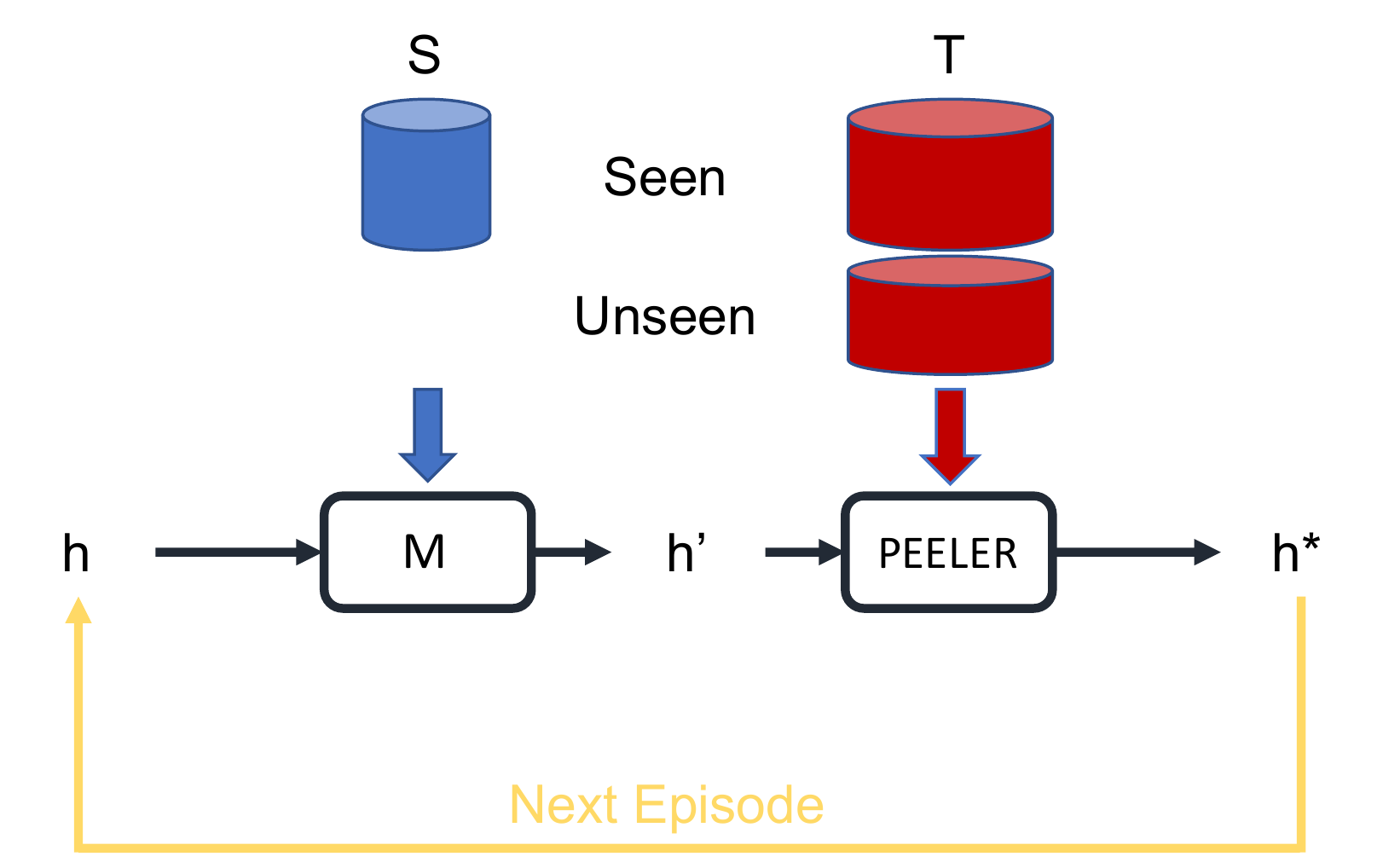}
	}
	\caption{The proposed general framework for open-set meta-learning: a meta training set $\{\mathbb{S}, \mathbb{T}\}$ is sampled, where $\mathbb{T}$ contains classes not in $\mathbb{S}$ as ``unseen'' classes, and loss in (\ref{eq:MLOS}) is minimized to obtain $h^*$.}
	\label{fig:osml}
  \vspace{-1em}
\end{figure}

\section{Meta Learning-based open-set recognition}

In this section, we introduce the proposed ML approach to open-set recognition. We first introduce the general procedure and then discuss the specific embedding metric used in our PEELER implementation.

\subsection{Open-Set Meta-Learning} 
As shown in Figure~\ref{fig:osml}, open-set meta-learning (PEELER) relies on a meta training set $\mathbb{MS} = \{(\mathbb{S}^s_i, \mathbb{T}^s_i)\}_{i=1}^{N^s}$, and a meta test set $\mathbb{MT} = \{(\mathbb{S}^t_i, \mathbb{T}^t_i)\}_{i=1}^{N^t}$. The only difference with respect to standard ML is that the episodes $(\mathbb{S}, \mathbb{T})$ are open set. While 
the training set $\mathbb{S}$ is identical to the one used in standard ML, the test set $\mathbb{T}$ is {\it augmented with unseen classes\/}. Hence, PEELER can be addressed with a solution similar to the ML procedure of Section~\ref{sec:ML}. While the ML step remains as in~(\ref{eq:ML1}), the optimization step of~(\ref{eq:ML2}) becomes
\begin{eqnarray}
    h^* &=& \arg \min_{h} \left\{\sum_{(x_k, y_k) \in \mathbb{T}^s_i | y_k \in \mathbb{C}_i^s} L_c[y_k, h'(x_k)]
    \right. \nonumber \\
    &+& \left. \lambda \sum_{(x_k, y_k) \in \mathbb{T}^s_i | y_k \in \mathbb{C}_i^u} L_o[h'(x_k)]\right\}
    \label{eq:MLOS}
\end{eqnarray}
where $\mathbb{C}_i^s$ ($\mathbb{C}_i^u$) is the set of seen (unseen) classes of $\mathbb{T}^s_i$, $L_c[.,.]$ is the classification loss to apply to the seen classes (typically the cross-entropy loss) and $L_o$ is an open-set loss applied to those unseen in $\mathbb{S}^s_i$.

\parsp
\noindent{\bf Few-shot open-set recognition.}
The few-shot setting requires the sampling of the classes that make up the support and query sets. Similarly to closed-set few-shot recognition, the support set $\mathbb{S}^s_i$ of the $i^{th}$ episode is obtained by sampling $N$ classes and $K$-examples per class. This defines the seen classes $\mathbb{C}_i^s$. However, the query set $\mathbb{T}^s_i$ is composed by the combination of these classes with $M$ additional unseen classes $\mathbb{C}_i^u$. These support and query sets are used in (\ref{eq:MLOS}).

\parsp
\noindent{\bf Large-scale open-set recognition.} In the large-scale setting, the seen classes have plenty of examples and are well trained without any need for resampling. 
% In this case, all classes are included in $\mathbb{S}^s_i$, which is constant [CHECK]. 
However, resampling of the unseen classes is still advantageous, since it enables an embedding that generalizes better to these classes.
In each episode, $M$ classes are randomly sampled  from the class label space to form the set of unseen classes $\mathbb{C}_i^u$, and the remaining classes are used as seen classes $\mathbb{C}_i^s$. 
Without a support set $\mathbb{S}^s_i$, the meta-learning step of~(\ref{eq:ML1}) is no longer needed. Instead, we rely on the set of seen classes to adjust the model to only classify samples into those classes, {\it i.e.} a mapping
\begin{equation}
    h' = {\cal M}(h,\mathbb{C}_i^s),
    \label{eq:ML1_LARGE}
\end{equation}
with the loss function of~(\ref{eq:MLOS}) is still applied.

\parsp
\noindent{\bf Open-set loss.} During inference, when faced with samples from unseen classes, the model should not assign a large probability to any class. In this case, a sample can be rejected if the largest class probability $\max_k p_\phi (y = k | \mathbf{x})$ among seen classes is small. To enable this, the learning algorithm should minimize the probabilities on seen classes for samples from $\mathbb{C}_i^u$. This can be implemented by maximizing the entropy of seen class probabilities, i.e  using the negative entropy
\begin{equation}
    L_o[\mathbf{x}] = \sum_{k\in \mathbb{C}_i^s} p(y = k | \mathbf{x}) \log p(y = k | \mathbf{x})
\end{equation}
as loss function. 

\begin{figure*}[t!]
\centering{
		\begin{tabular}{cccc} 
		\includegraphics[width=0.23\textwidth]{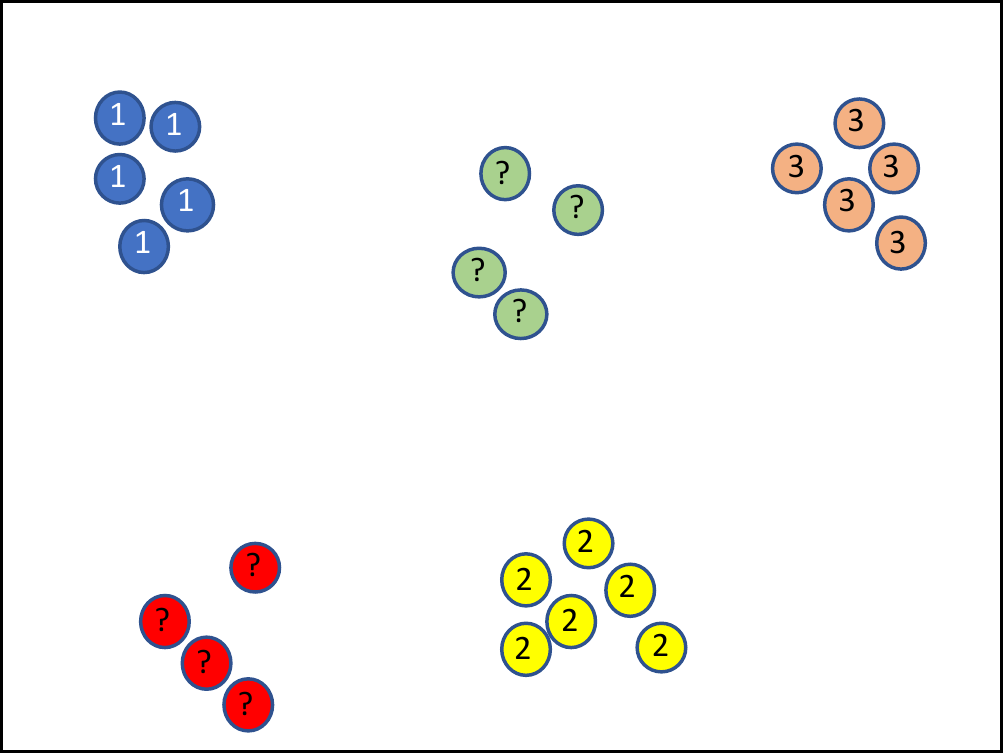} & \includegraphics[width=0.23\textwidth]{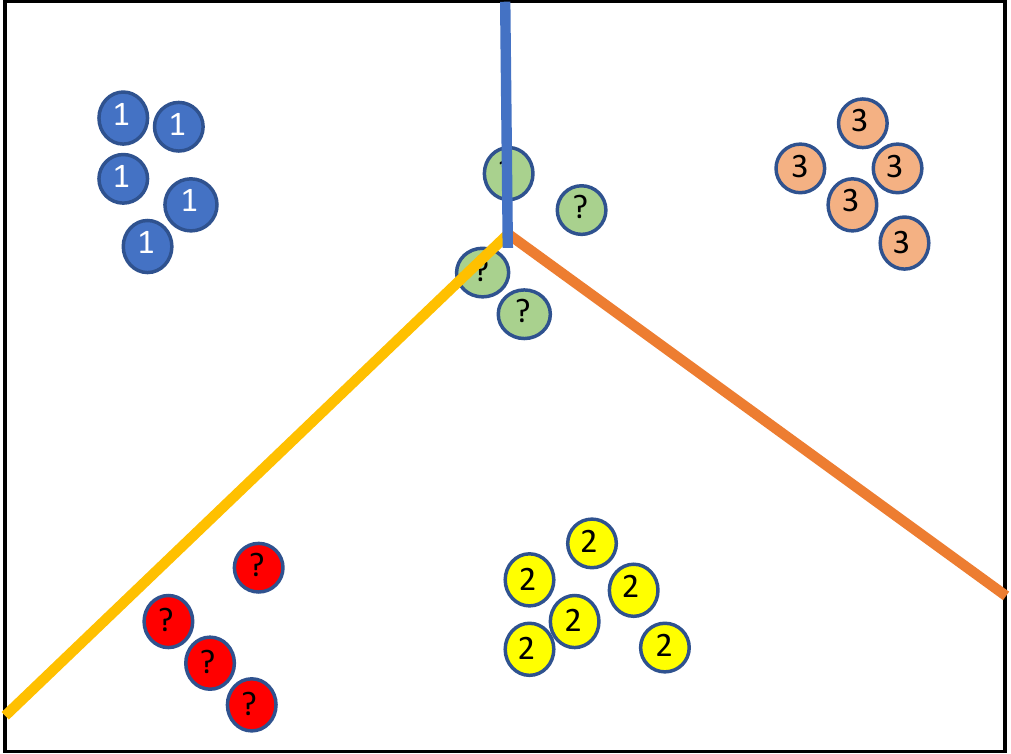} &
		\includegraphics[width=0.23\textwidth]{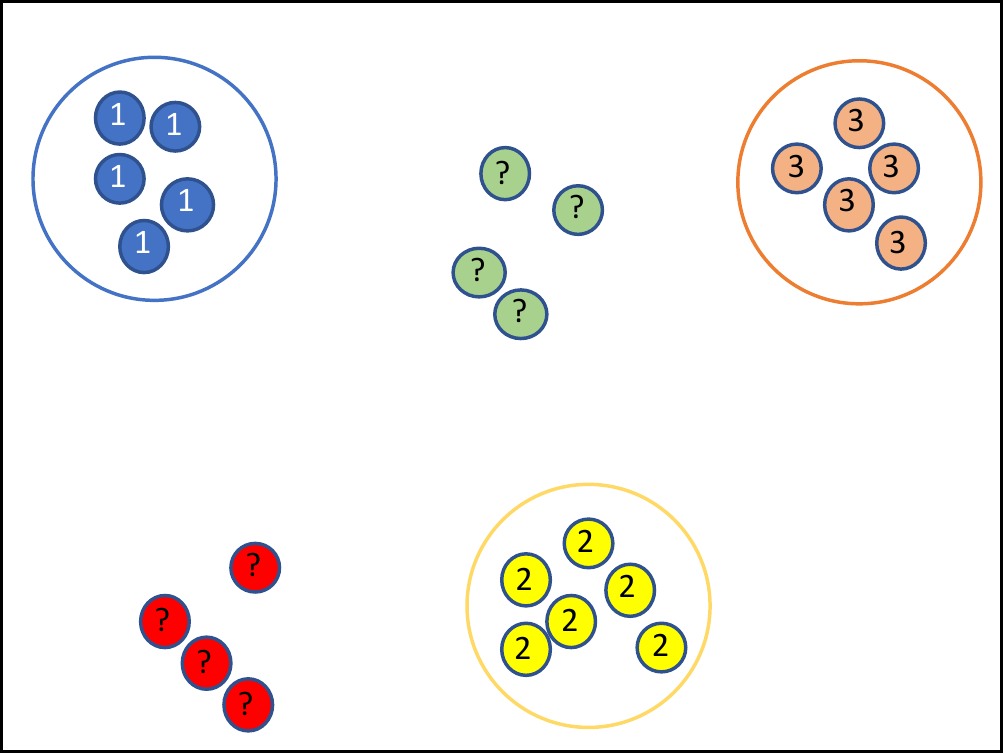} &
		\includegraphics[width=0.23\textwidth]{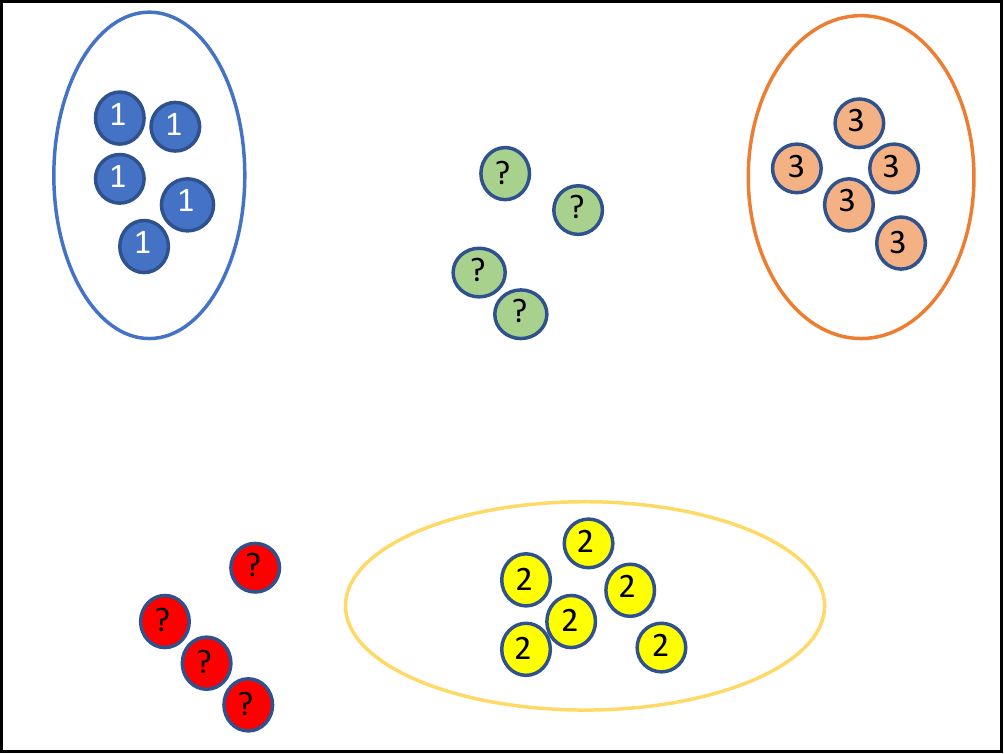} \\
		(a) & (b) & (c) & (d)
		\end{tabular}}
	\caption{(a) Feature space of an open-set recognition problem with 3 seen and 2 unseen classes; (b) Optimal boundaries for closed-set classification; (c) Each class defines an euclidean distance of equal radius. While optimal for closed-set, this is sub-optimal for open set recognition; (d) a better set of open-set distances can be learned by allowing different Gaussian clusters of class-dependent variance along each dimension. }
	\label{fig:openset}
  \vspace{-1em}
\end{figure*}

\subsection{Gaussian Embedding~\label{sec:GaussianE}} 
% We start by reviewing a well-known metric based few-shot method Prototypical Networks~\cite{Snell2017}, which combined metric learning with widely used cross-entropy loss and generalize it to both traditional and few-shot open-set recognition problems. More specifically, for each class $k$, prototypical networks assume a prototype $\mathbf{\mu}_k$ as a class center in a feature space which is defined by an embedding function $f_\phi$ with learnable parameters $\phi$. And the distance from a given feature $f_\phi (\mathbf{x})$ to class centers are measured as $d(f_\phi(\mathbf{x}), \mathbf{\mu}_k)$. During training, it produces a distribution of probability defined on the softmax over negative distances, and the negative log-likelihood is minimized as in (\ref{eq:softmaxdist}) and (\ref{eq:softmaxce}). In few-shot learning, each prototype is generated by the mean vector of the features embedded from support samples in the corresponding class:
% \begin{equation} \label{eq:proto}
%     \mathbf{\mu}_k = \frac{1}{|S_k|}\sum_{(\mathbf{x}_i, y_i)\in S_k}f_\phi (\mathbf{x}_i)
% \end{equation}
% , where $S_k$ is a set of all support samples from class $k$.
While PEELER is a general framework, in this work we propose an implementation based on the prototypical network architecture~\cite{Snell2017}. A set of class prototypes is first defined and samples with a large distance to the closest class prototype are assigned to the set of unseen classes. For low-shot classification, the class prototypes are the class means of (\ref{eq:proto}). For large-scale classification, we assume fixed prototypes that are embedded in the network and learned by back-propagation. %, shown in Figure~\ref{fig:model}.

Although several distances are discussed in~\cite{Snell2017}, the prototypical network is usually implemented with  the Euclidean distance, {\em i.e.},
% \begin{eqnarray}
%     d(f_\phi(\mathbf{x}), \mathbf{\mu}_k) &=& ||f_\phi(\mathbf{x}) - \mathbf{\mu}_k||^2 \\ \nonumber
%     &=& [f_\phi(\mathbf{x}) - \mathbf{\mu}_k]^T[f_\phi(\mathbf{x}) - \mathbf{\mu}_k]
% \end{eqnarray}
\begin{equation}
    d(f_\phi(\mathbf{x}), \mathbf{\mu}_k) = (f_\phi(\mathbf{x}) - \mathbf{\mu}_k)^T(f_\phi(\mathbf{x}) - \mathbf{\mu}_k).
    \label{eq:euclid}
\end{equation}
This implies that the features from each class follow a Gaussian distribution with mean $\mathbf{c}_k$ and diagonal covariance $\sigma^2I$, where $\sigma$ is shared by all classes. 
%However, this assumes a powerful embedding function, such that all samples are embedded into the feature space as expected. 
% In other word, we rely on the ability of feature extractor to reduce intra-class variation. 
While sensible for closed-set few-shot learning, where the embedding is learned to produced such feature distributions,
%as long as it is powerful enough. In fact, it has been show that performance improves for deeper feature extractors~\cite{Chen2019}. It 
this can be very sub-optimal when open-set samples are introduced. 
Figure~\ref{fig:openset} illustrates the problem for a setting with three seen classes and two unseen classes. Even though, as shown in Figure~\ref{fig:openset} (a), the embedding is such that seen classes are spherically distributed and have the same covariance, the open-set samples, unseen during training, are still embedded onto the feature space at random locations. Hence, although the optimal boundaries of the closed-set classifier, shown in  Figure~\ref{fig:openset} (b), match the contours of the distance of~\eqref{eq:euclid}, shown in  Figure~\ref{fig:openset} (c), they are not optimal for open set recognition. In fact, as shown in Figure~\ref{fig:openset} (d), the shape of the optimal boundary between seen and unseen classes can even vary from one class to another. 

To account for this, we assume a Gaussian distribution of mean $\mathbf{\mu}_k$ and covariance $\Sigma_k$ for class $k$. The Euclidean distance of~\eqref{eq:euclid} is thus replaced by  the Mahalanobis distance
\begin{equation}
    d(f_\phi(\mathbf{x}), \mathbf{\mu}_k) = [f_\phi(\mathbf{x}) - \mathbf{\mu}_k]^T\Sigma_k^{-1}[f_\phi(\mathbf{x}) - \mathbf{\mu}_k].
\end{equation}
To keep the number of parameters manageable, we assume that all covariance matrices are diagonal, {\it i.e.} $\Sigma_k = \text{diag}(\sigma_{k1}, \sigma_{k2}, \ldots, \sigma_{kM})$, and the precision matrix $A_k=\Sigma_k^{-1}$ is used to ease calculations.
%{\it i.e.}
% \begin{eqnarray}
%     \Sigma_k &=& \text{diag}(\sigma_{k1}, \sigma_{k2}, \ldots, \sigma_{kM}) \\
%     A_k &=& \text{diag}(\sigma_{k1}^{-1}, \sigma_{k2}^{-1}, \ldots, \sigma_{kM}^{-1})
% \end{eqnarray}
Similar to the class prototypes, the learning of precision matrices depends on the recognition setting. For large-scale open-set recognition, the $A_k$ are network parameters learned directly by back-propagation.
In the few-shot setting, where support samples are available, we introduce a new embedding function $g_\varphi$ with learnable parameters $\varphi$, and define
\begin{equation} \label{eq:weight}
    A_k = \frac{1}{|S_k|}\sum_{(\mathbf{x}_i, y_i)\in S_k}g_\varphi (\mathbf{x}_i).
\end{equation}
\section{Experiments}

The proposed method was compared to state-of-the-art (SOTA) approached to open-set recognition. Following Neal et al.~\cite{Neal2018}, we evaluate both classification accuracy and open-set detection performance. To clarify the terminology, closed-set classes are the categories seen during training and open-set classes the novel categories only used for testing. %Not all samples from the closed-set classes would be used for training that there are both training and testing samples in the closed-set classes. 
All training is based on the training samples from closed-set classes. For testing, we use the testing samples from both the training and open-set classes. Classification {\it accuracy} is used to measure how well the model classifies closed-set samples, i.e., test samples from closed-set classes. The {\it AUROC} (Area Under ROC Curve) metric is used to measure how well the model detects open-set samples, i.e., test samples from open-set classes, within all test samples. To simplify the writing, we define the following acronyms: {\it our basic} represents prototypical networks with euclidean distance for open-set detection; {\it GaussianE} represents the Gaussian Embedding introduced in Sec.~\ref{sec:GaussianE}; {\it OpLoss} represents the proposed open-set loss.

\subsection{Large-Scale Open-Set Recognition}

Most prior works on open-set recognition are designed for the large-scale setting. Here we evaluate PEELER on CIFAR10~\cite{krizhevsky2014cifar} and extended miniImageNet~\cite{Gidaris2018}.
CIFAR10 consists of $60,000$ images of $10$ classes. Following~\cite{Neal2018}, 6 classes are first randomly selected as closed-set classes, and the other 4 kept as open-set classes. Results are averaged over 5 random partitions of closed/open-set classes. %The training/testing split is adopted from the original dataset. 
Extended miniImageNet is designed for few-shot learning. We use the $64$ training categories and $600$ images per category as closed set training data, while the $300$ extra images of each category are used for closed set testing. Images from $20$ test categories are used for open set testing.

\parsp
\noindent{\bf Training.}
On CIFAR10, we randomly sample 2 from the 6 closed-set classes per episode to apply the open-set loss, the remaining 4 classes are used to train classification. On miniImageNet, the closed/open-set partition is fixed. We use Adam~\cite{kingma2014adam} with an initial learning rate of $0.001$, $\lambda = 0.5$ in (\ref{eq:MLOS}) and $10,000$ training episodes. The learning rate is decayed by a factor of $0.1$ after $6,000$ and $8,000$ episodes.

\begin{table}[t!]
	\begin{footnotesize}
		\centering
		\begin{tabular}{c|cc}
		    \hline
			Model & Accuracy(\%) & AUROC(\%) \\
		    \hline
		    \multicolumn{3}{c}{ConvNet on CIFAR10} \\
			\hline
			Softmax & 80.1 & 67.7 \\
			OpenMax & 80.1 & 69.5 \\
			G-OpenMax & 81.6 & 67.9 \\
			Counter & 82.1 & 69.9 \\
			\hline
			Confidence & 82.4 & 73.19 \\
			\hline
			Our basic & 82.4 & 74.62 \\
			Our basic + GaussianE & 82.3 & 75.65 \\
			Our basic + GaussianE + OpLoss & 82.3 & {\bf 77.22} \\
			\hline
		    \hline
		    \multicolumn{3}{c}{ResNet18 on CIFAR10} \\
			\hline
			Softmax & 94.2 & 78.90 \\
			OpenMax & 94.2 & 79.02 \\
			\hline
			Confidence & 94.0 & 80.90 \\
			\hline
			Our basic & 94.7 & 82.94 \\
			Our basic + GaussianE & 94.3 & 83.12 \\
			Our basic + GaussianE + OpLoss & 94.4 & {\bf 83.99} \\
			\hline
		    \hline
		    \multicolumn{3}{c}{ResNet18 on miniImageNet} \\
			\hline
			Softmax & 76.1 & 76.65 \\
			OpenMax & 76.1 & 77.80 \\
			\hline
			Confidence & 76.5 & 80.67 \\
			\hline
			Our basic & 76.4 & 80.59 \\
			Our basic + GaussianE & 76.1 & 81.06 \\
			Our basic + GaussianE + OpLoss & 76.3 & {\bf 82.12} \\
			\hline
		\end{tabular}
		\caption{Comparison to SOTAs on large-scale open-set recognition: PEELER outperforms all others in terms of open-set sample detection {\it AUROC},  for a comparable classification accuracy.}
		\label{tab:many_results}
	\end{footnotesize}
  \vspace{-1em}
\end{table}

\parsp
\noindent{\bf Results.}
We compare the proposed method to several open-set recognition SOTA methods, including OpenMax~\cite{Bendale2016}, G-OpenMax~\cite{Ge2019}, and Counterfactual~\cite{Neal2018}, and an out-of-distribution SOTA method, Confidence~\cite{lee2017training}. All models use the same CNN backbone for fair comparison. We tested both a CNN, denoted as ConvNet, proposed by~\cite{Neal2018}, and ResNet18~\cite{he2016deep}.

Table~\ref{tab:many_results} shows that, for both backbones, PEELER outperforms all previous approaches by a significant margin. For a similar classification accuracy on seen classes, it achieves much higher AUROCs for the detection of unseen classes. The open-set detection performance is enhanced by both the proposed Gaussian embedding and open-set loss. 
%While compared to previous methods all of our work improves the open-set detection by a large margin on ConvNet, the improvement is not as large when it comes to ResNet18. One possible reason is that the features from ResNet are already close to Gaussian distribution.

\subsection{Few-Shot Open-Set Recognition} \label{sec:fsos}
\parsp
\noindent{\bf Dataset.}
Few-Shot Open-Set performance is evaluated on mini-Imagenet~\cite{Vinyals2016}, using the splits of~\cite{Ravi2017}. 64 classes are used for training, 16 for validation and another 20 for test.

\parsp
\noindent{\bf Training.}
The open-set problem follows the 5-way few-shot recognition setting. During training, $10$ classes are randomly selected from the training set per episode, $5$ are used as closed-set classes and the other $5$ as unseen. All support set samples are from the closed-set classes. The query set contains samples from closed-set classes, the closed-query set, and samples from open-set classes, the open-query set.
%closed-set classes and open-set classes are randomly selected from training set.
%$5$ classes are sampled from the training set as closed-set classes. Query set consists of samples from
%During training, 5 classes are first sampled from the training set. Support and closed-query sets are then assembled from those 5 classes. The open-query set is collected from another $5$ classes in the training set. Both closed and open-query sets have $15$ samples per class. The combination of closed and open-query sets is used to compute the loss in (\ref{eq:MLOS}), where $\lambda = 0.5$. 
The evaluation set is sampled from the test set with the same strategy. The evaluation is repeated $600$ times to minimize uncertainty. The total number of training episodes is $30,000$. The learning rate is decreased by a factor of $0.1$ after $10,000$ and $20,000$ episodes.

\begin{table}[t!]
	\begin{footnotesize}
		\centering
		\begin{tabular}{c|cc}
		    \hline
			Model & Accuracy(\%) & AUROC(\%) \\
		    \hline
		    \multicolumn{3}{c}{5-way 1-shot} \\
			\hline
			GaussianE + OpenMax & 57.89$\pm$0.59 & 58.92$\pm$0.59 \\
			GaussianE + Counterfactual & 57.89$\pm$0.59 & 52.20$\pm$0.61 \\
			\hline
			Our basic & 56.31$\pm$0.57 & 58.94$\pm$0.60 \\
			Our basic + OpLoss & 56.34$\pm$0.57 & 60.94$\pm$0.61 \\
			Our basic + GaussianE & 57.89$\pm$0.59 & 58.66$\pm$0.60 \\
			Our basic + GaussianE + OpLoss & {\bf 58.31$\pm$0.58} & {\bf 61.66$\pm$0.62} \\
		    \hline
		    \hline
		    \multicolumn{3}{c}{5-way 5-shot} \\
			\hline
			GaussianE + OpenMax & 75.31$\pm$0.76 & 67.54$\pm$0.67 \\
			GaussianE + Counterfactual & 75.31$\pm$0.76 & 53.25$\pm$0.59 \\
			\hline
			Our basic & 74.19$\pm$0.75 & 66.00$\pm$0.67 \\
			Our basic + OpLoss & 74.14$\pm$0.74 & 67.92$\pm$0.68 \\
			Our basic + GaussianE & {\bf 75.31$\pm$0.76} & 66.50$\pm$0.67 \\
			Our basic + GaussianE + OpLoss & 75.08$\pm$0.72 & {\bf 69.85$\pm$0.70} \\
			\hline
		\end{tabular}
		\caption{Few-shot open-set recognition results. Comparison to several baselines and prior open-set methods. 
		%Compared with prototypical networks as a baseline. ``Proto+Entropy'' means apply entropy loss on Euclidean distance. And ``Gauss'' means use Gaussian embedding.
		\label{tab:few_results}}
	\end{footnotesize}
  \vspace{-1em}
\end{table}

\begin{table}[t!]
	\begin{footnotesize}
		\centering
		\begin{tabular}{c|cc}
		    \hline
			Model & Accuracy(\%) & AUROC(\%) \\
			\hline
			Our basic & 39.61$\pm$0.40 & 71.32$\pm$0.70 \\
			Proto+Entropy & 39.41$\pm$0.40 & 72.23$\pm$0.72 \\
			Our basic + GaussianE & 40.18$\pm$0.40 & 71.31$\pm$0.70 \\
			Our basic + GaussianE + OpLoss & {\bf 41.90$\pm$0.39} & {\bf 74.97$\pm$0.74} \\
		    \hline
		\end{tabular}
		\caption{10-way 1-shot openset recognition results. The AUROC is higher than those in 5-way.
		\label{tab:few_results_diff}}
	\end{footnotesize}
  \vspace{-1.5em}
\end{table}

\parsp
\noindent{\bf Results.}
Since not all prior open-set methods support the few-shot setting, some modifications were required. For example, generative methods~\cite{Ge2019, Neal2018} do not support few-shot samples. Instead, we train the model on the pre-training set and fine-tune it on the support set.
The closed-set classifier is as above, and a two-class classifier is further trained to detect open-set samples. OpenMax~\cite{Bendale2016} is easier to apply under the few-shot setting. We apply OpenMax on the activations of the pre-softmax layer, {\it i.e.} the negative of the distance of our Gaussian setting. 
All methods are implemented with the ResNet18~\cite{he2016deep} for fair comparison. 

% In addition to our methods, we also tried to implement regular open-set models on few-shot learning. However, due to the difference in the training stage, it is not obvious for some methods to generalize to the few-shot scenario. In fact, only the OpenMax~\cite{Bendale2016} can be applied easily. In the following experiments, we apply OpenMax on the activations before softmax layer, {\it i.e.} the negative distance in our Gaussian setting. Generative open-set models~\cite{Ge2019, Neal2018} cannot be trained with few-shot samples. However, to show how bad they are, we choose to implement~\cite{Neal2018} as an example. Actually, in the implementation, the generator is only trained on the pre-training set. We tried to fine-tune it on the few-shot support set, but it totally failed. The closed-set classifier is used the same as our Gaussian setting, and a two-class classifier is further trained with to detect open-set samples. All methods are implemented with the same ResNet18~\cite{he2016deep} structure to ensure the fair comparison. 

As shown in Table~\ref{tab:few_results}, both OpenMax and Counterfactual have weaker performance than the proposed approach. Note that a {\it AUROC} of $50\%$ corresponds to chance performance. The proposed open-set loss, Gaussian embedding, and their combination all provide gains.
% The result close to $50\%$ means a total failure of detecting unseen classes. Although many prototypical networks results are reported in the literature, and some of them are based on ResNet. We have to reproduce it by ourselves because we need to evaluate on the open-set problem. And in the sense of few-shot accuracy, our implementation is good enough compared to some of the best available implementations~\cite{Chen2019,Li2019}. 
% When the covariance predictor $g_\varphi$ is applied, it shares the first three residue blocks with $f_\phi$, and has a separate fourth residue block. As both the number of samples and the number of classes in closed-query and open-query sets are the same, one can expect a $50\%$ if detecting open samples randomly. However, from what we get from prototypical networks, although the method works good in closed-set recognition, it is only slightly better than chance when dealing with open-set recognition in few-shot learning. When we add Gaussian variance, the classification accuracy improves simultaneously. When we add entropy loss directly on the prototypical network, the AUROC for open-set recognition improves but the classification accuracy doesn't change. Finally, when both of them are applied, classification and open-set performance improve at the same time even larger than any of each separate. This implies that the Gaussian assumption on each class prototype is better to build a feature space. And our entropy loss is effective to train a better open-set model, while maintains the closed-set accuracy.
\begin{figure*}[t!]
\centering{
		\begin{tabular}{ccc} 
		\includegraphics[width=0.28\textwidth]{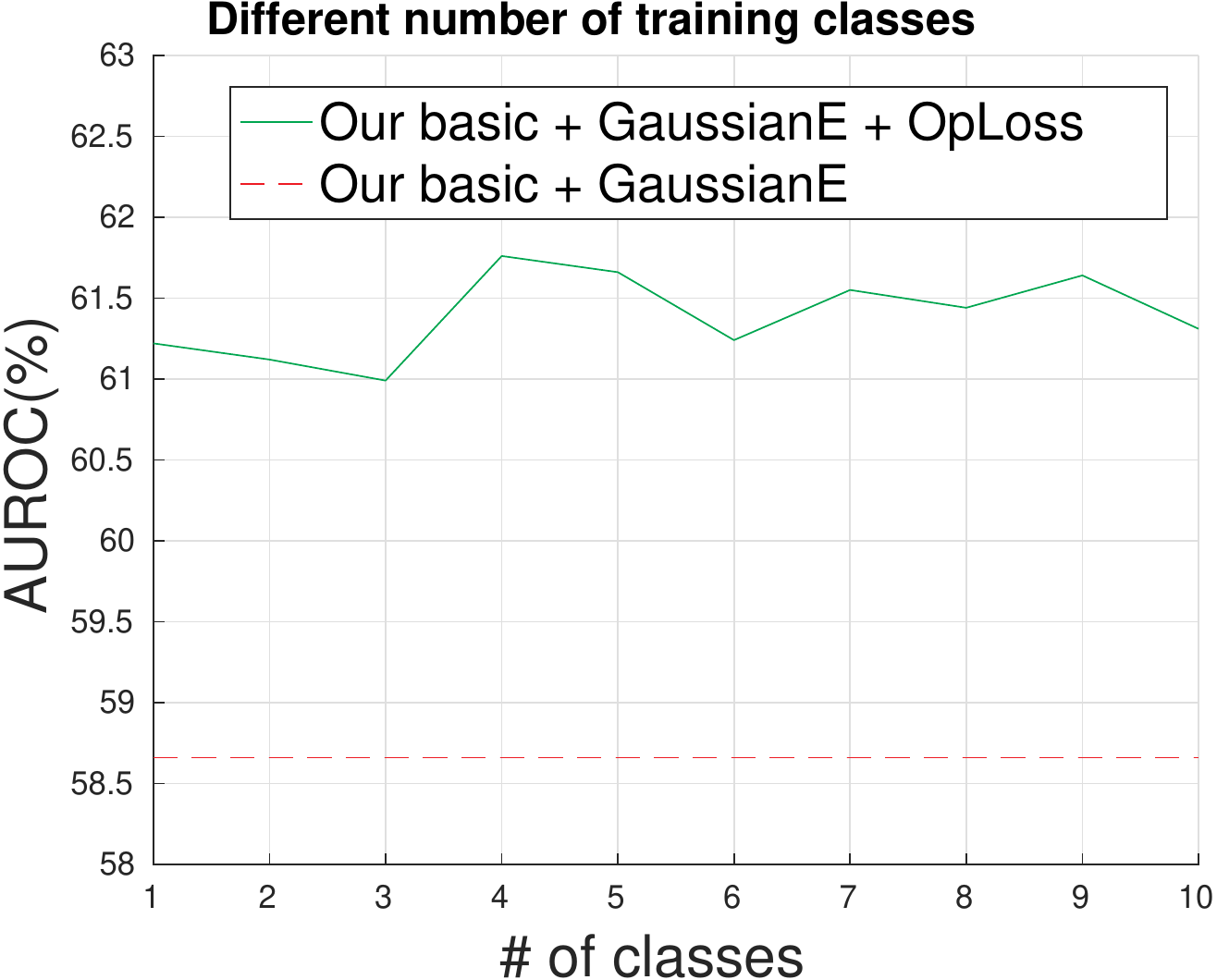} & \includegraphics[width=0.28\textwidth]{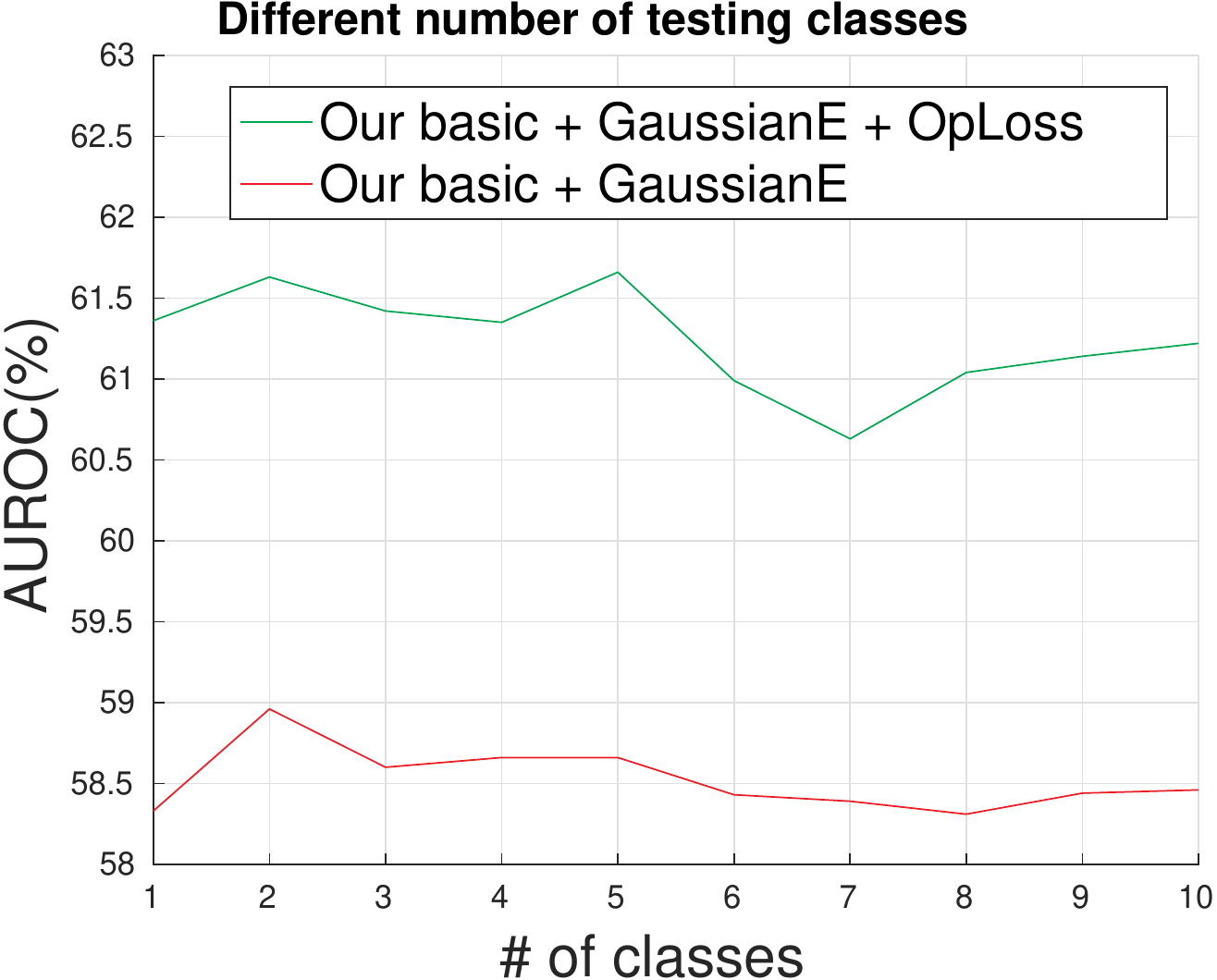} &
		\includegraphics[width=0.28\textwidth]{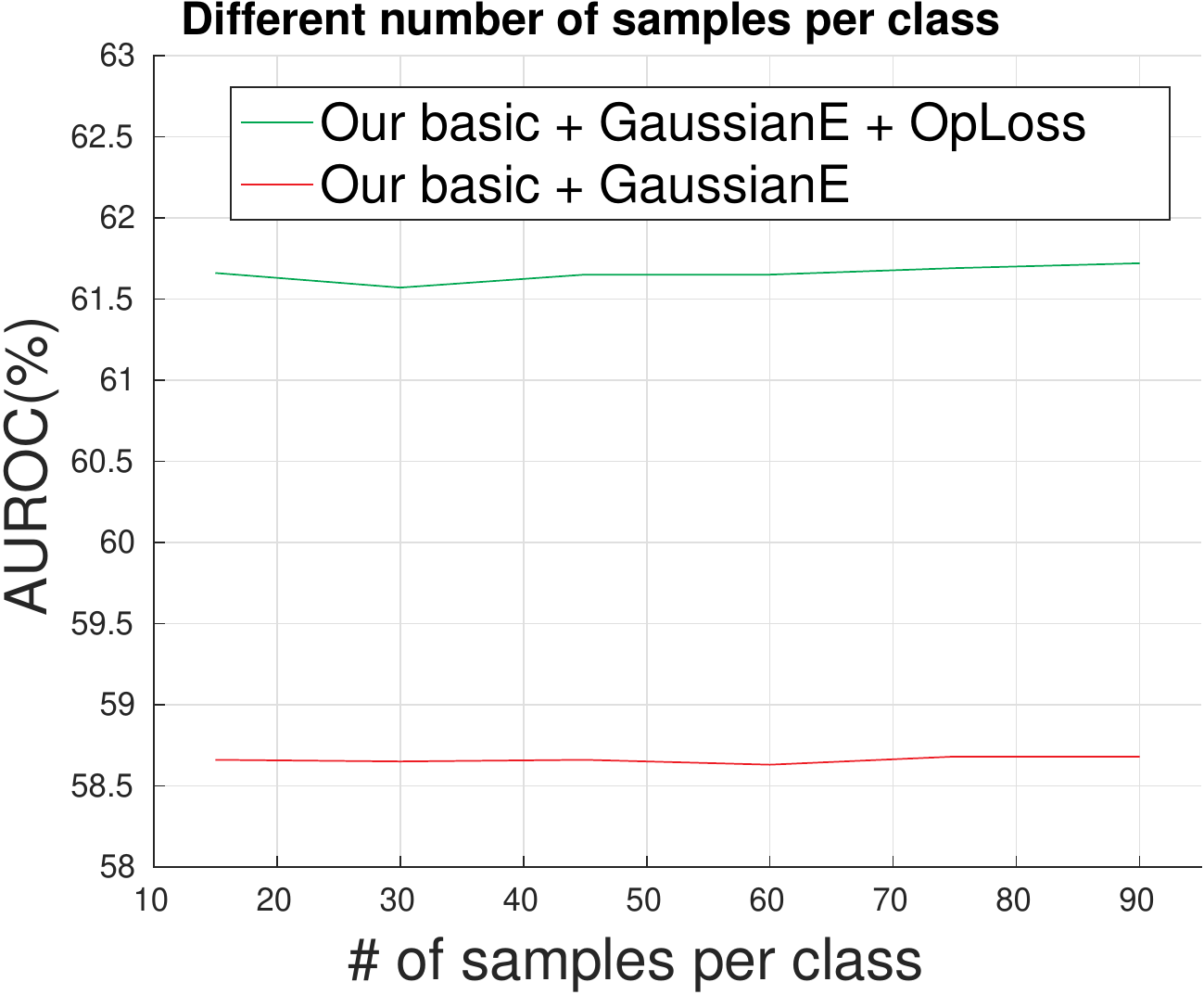} \\
		(a) & (b) & (c)
		\end{tabular}}
	\caption{Ablation Study: the proposed open-set loss produces consistent gain with different (a) number of training classes, (b) number of testing classes, or (c) number of testing samples per class.}
	\label{fig:different_results}
  \vspace{-1em}
\end{figure*}

\subsection{Ablation Studies}
%In the following experiments, the closed-query set keeps the same as that in the previous section, while the open-query set is modified. 

\paragraph{Sampling Strategy in Training.}
We study how the training sampling strategy affects open-set detection results. The number of unseen classes used to produce open-set samples by training episode is varied, for a fixed total number of open-set samples. The corresponding open-set detection results are shown in Figure~\ref{fig:different_results} (a). The performance variation is minor, when compared to the average gain over the baseline method. We hypothesize that this robustness is due to task randomness. When the total number of training episodes is large, the model converges well no matter how many open-set classes are included in a single episode.

\vspace{-1em}
\paragraph{Factors in Open-set Testing.}
For large-scale open-set recognition, the number of open-set samples follows from the number of open-set classes. The difficulty of the open-set problem is determined by the latter. We try to discover factors that determine the difficulty of the few-shot open-set problem. A first factor is the number of classes in the open-query set. We vary this number from 1 to 10, while keeping the training procedure and the total number of samples in the open-query set unchanged. Results are shown in Figure~\ref{fig:different_results} (b). 
A second factor is the number of samples per class, which is changed while the number of classes remains $5$. Results are shown in Figure~\ref{fig:different_results} (c).
The figures show that variations cause small changes in AUROC performance. This means that the factors have little impact in the difficulty of the problem.

\begin{table}[t!]
	\begin{footnotesize}
		\centering
		\begin{tabular}{lccc|ccc}
		    \hline
			\multirow{2}{*}{Category} & \multicolumn{3}{c|}{VODC~\cite{wang2016video}} & \multicolumn{3}{c}{Ours} \\
			\cline{2-7}
			& I & II & III & I & II & III \\
		    \hline
		    airplane & 20 & 10 & {\bf 0} & 6 & 6 & {\bf 0} \\
		    balloon & 13 & 4 & 3 & 11 & 7 & {\bf 2} \\
		    bear & 3 & 3 & {\bf 2} & {\bf 2} & {\bf 2} & {\bf 2} \\
		    cat & 4 & 5 & 5 & 5 & {\bf 3} & {\bf 3} \\
		    eagle & 23 & 12 & 8 & 5 & 2 & {\bf 0} \\
		    ferrari & 11 & 7 & 6 & 11 & {\bf 3} & {\bf 3} \\
		    figure skating & {\bf 0} & {\bf 0} & {\bf 0} & {\bf 0} & {\bf 0} & {\bf 0} \\
		    horse & 5 & {\bf 1} & {\bf 1} & 5 & 4 & 3 \\
		    parachute & 14 & 10 & 2 & 10 & 2 & {\bf 0} \\
		    single diving & 18 & 13 & 5 & 4 & 2 & {\bf 1}  \\
			\hline
			{\bf Avg.} & 11.1 & 6.5 & 3.2 & 5.9 & 3.1 & {\bf 1.4} \\
			\hline
		\end{tabular}
		\caption{The number of misclassified frames (lower is better) by varying the number of annotated frames (I, II, III in the second row).
		\label{tab:xs}}
	\end{footnotesize}
  \vspace{-1em}
\end{table}

\vspace{-1em}
\paragraph{Factors in Few-Shot Testing.}
We compare 5-way to 10-way classification, for a fixed open-set component of both training and testing sets. Results are shown in Table~\ref{tab:few_results_diff}. Although, as expected,  10-way underperforms 5-way classification, the open-set performance improves significantly. This shows that few-shot classification tasks with more class diversity are more robust to open-set samples.

% \subsection{Feature Visualization~\textcolor{red}{MORE EXPERIMENTS}}
% Figure~\ref{...} visualize the feature distribution from t-SNE~\cite{...}. 

\subsection{Weakly Supervised Object Discovery}
Finally, we investigate an application of few-shot open-set recognition. A weakly supervised object discovery problem is considered in~\cite{wang2016video}. A video of an object is given, but some frames are irrelevant for the presence of the object. The task is to find these irrelevant frames when the number of annotated frames (object presented or not) is limited. For the open-set problem, only relevant frame labels are given, {\it i.e.} frames containing the object. The irrelevant frames are detected as open-set samples.

% \parsp
% \noindent{\bf Dataset:} 
The XJTU-Stevens Dataset~\cite{wang2016video} is adopted for evaluation. It has $101$ videos from $10$ categories with different instances, and some frames in those videos do not have the object. During training, a 5-way 1-shot few-shot open-set model is trained as described in Sec.~\ref{sec:fsos}. Instead of category level classification, we perform instance level classification of each frame. During testing, only one video is considered, using the frames annotated as relevant  as the support set. This implies that only one Gaussian class center is provided. The unlabeled frames are labeled as seen or unseen by their Mahalanobis distance to the center. Frames with distance higher than a threshold are detected as irrelevant. The number of annotated relevant frames varies from 1 to 3. Results are shown as the number of misclassified frames. The best  VODC method in~\cite{wang2016video} is listed for comparison. Note that VODC requires the same number of annotated irrelevant frames, which PEELER does not. The proposed method largely outperforms VODC, halving the number of misclassified frames in all three settings. This shows that its feature embedding is a better solution for open-set detection.

% \begin{table}[t!]
% 	\begin{footnotesize}
% 		\centering
% 		\begin{tabular}{c|cc}
% 		    \hline
% 			Model & Accuracy & AUROC \\
% 		    \hline
% 		    \multicolumn{3}{c}{5-way 1-shot} \\
% 			\hline
% 			Gauss & 94.25$\pm$0.95 & 89.82$\pm$0.90 \\
% 			%Gauss+OpenMax & & \\
% 			Gauss+Entropy & 94.07$\pm$0.95 & 93.34$\pm$0.94 \\
% 			\hline
% 		    \hline
% 		    \multicolumn{3}{c}{5-way 5-shot} \\
% 			\hline
% 			Gauss & 97.87$\pm$0.98& 93.59$\pm$0.93\\
% 			%Gauss+OpenMax & & \\
% 			Gauss+Entropy & 98.21$\pm$0.98& 96.89$\pm$0.97\\
% 			\hline
% 		    \hline
% 		    \multicolumn{3}{c}{20-way 1-shot} \\
% 			\hline
% 			Gauss & 88.02$\pm$0.88& 88.22$\pm$0.88\\
% 			%Gauss+OpenMax & & \\
% 			Gauss+Entropy & 85.98$\pm$0.86& 88.31$\pm$0.88\\
% 			\hline
% 		    \hline
% 		    \multicolumn{3}{c}{20-way 5-shot} \\
% 			\hline
% 			Gauss & 94.09$\pm$0.94 & 92.70$\pm$0.93\\
% 			%Gauss+OpenMax & & \\
% 			Gauss+Entropy & 93.58$\pm$0.93 & 93.62$\pm$0.93 \\
% 			\hline
% 		\end{tabular}
% 		\caption{Few-shot object discovery results.
% 		\label{tab:xs}}
% 	\end{footnotesize}
% \end{table}

%------------------------------------------------------------------------
\section{Conclusion}
In this work, we have revisited the open-set recognition problem in the context of few-shot learning. We proposed an extension of meta-learning that includes an open-set loss, and a better metric learning design. The resulting classifiers provide a new stat-of-the-art for few-shot open-set recognition, on mini-Imagenet. It was shown that, with few modifications, the approach  can also be applied to large-scale recognition, where it outperforms state of the art methods for open-set recognition. Finally, the XJTU-Stevens Dataset was used to demonstrante the effectiveness of the proposed model on a weakly supervised object discovery task.

\noindent{\bf Acknowledgement.}
Gang Hua was supported partly by National Key R\&D Program of China Grant 2018AAA0101400 and NSFC Grant 61629301. Bo Liu and Nuno Vasconcelos were partially supported by NSF awards IIS-1637941, IIS-1924937, and NVIDIA GPU donations.

{\small
\bibliographystyle{ieee_fullname}
\bibliography{egbib}
}

\end{document}